# Fabric defect detection using tactile information

Xingming Long[1], Bin Fang[1]*, Yifan Zhang[2], GuoYi Luo[1], Fuchun Sun[1]

*Abstract*—**This paper proposes a method of fabric structure defect detection based on tactile information. Different from traditional visual-based detection methods, the proposed method uses a tactile sensor to attain the information of fabric. The advantage of using tactile information is to avoid different irregular dyeing patterns and reduce the influence of ambient light. Therefore, the proposed method can be more concise and universal, which makes the defect detection system more robust. Experiments are conducted to verify the performance of the developed tactile method. The results demonstrate that the proposed method is much better than the visual method in the detection of structural defects. In addition, we propose a design of a tactile sensing device that can greatly improve the actual detection efficiency of the tactile method. It is showed that the proposed method is efficient enough to meet our requirements and it provides a possible solution for the application of the tactile method in actual fabric production.**

## I. INTRODUCTION

Fabric is one of the necessities which humans highly rely on in daily life. Clothing, blankets, and bandages are all made of fabric. Since there is a tremendous amount of demand for fabric, the textile industry flourishes to satisfy the demand through textile manufacturing. However, fabric defects will reduce the value of the fabric by about 45%-65% [1]. Therefore, fabric defect detection is indispensable in the production process for fabric quality control [2]. Traditional fabric defect detection by human eyes is tiring work and the quality of fabric can't be guaranteed due to fatigue [3]. Therefore, it is important to find automatic fabric defect detection methods with high robustness and accuracy, which can greatly improve the fabric production efficiency and reduce the labor cost.

Until now, most fabric defect detection methods are based on visual information. By setting up a camera above the fabric, the captured images are used to detect the defects. The fabric defect detection algorithms can be divided into three categories: statistical, spectral, and model-based methods. Statistical methods define various representations to describe the features of the fabric texture, which are used to distinguish the difference between the normal image and the image with defects. For example, [4] used histogram features, [5]-[7] analyzed the GLCM and LBP. [8][9] developed fractal method. Spectral methods realize the defect detection by transforming the input images to the spectral domain, including Fourier [10], Gabor [11], and wavelet [12][13] methods. The model-based approach uses modeling and decomposition methods [14]. Among all the model-based methods, the low-rank decomposition method is the most commonly used [15]-[17].

After the emergence of deep learning algorithms, they are widely mentioned and used in the field of artificial intelligence [18]-[20]. Among these deep learning algorithms, deep convolutional neural networks are more used in image processing for their powerful feature extraction ability. Various new convolution neural network models appear constantly with better performance on image processing, like GoogleNet [21], InceptionV3 [22], ResNet [23] and DenseNet [24]. Some researchers have tried fabric defect detection tasks using deep learning methods [25]-[27] and it is proved that the convolutional neural network can effectively extract features and detect the defects from fabric images.

In this paper, different from the traditional detection methods, we propose a method to detect fabric structure defects using tactile information. We divide the current defects of fabric into two categories: color defects and structural defects. Among them, color defects refer to the appearance defect caused by uneven dyeing or oil stain during production. Structure defects refer to the problems of flaws, knots, or even holes on the fabric during production. Compared with color defects, structure defects are more damaged and more difficult to identify. We use a tactile sensor to feel the fabric by touch and the sensor can get the tactile information from the fabric surface. Then, we can utilize tactile information to realize defect detection. The advantage of using tactile information for fabric defect detection is that the influence of dyeing patterns on the cloth can be avoided. These patterns are often annoying problems in the detection method using visual information, which will seriously affect the universality and complexity of the detection algorithm. Besides, the use of tactile information for defect detection can minimize the impact of the environment (such as the light condition), and effectively improve the robustness of the method.

The main contributions of this paper can be summarized in the following points:

(1) A new method of fabric defect detection using tactile information is proposed, which can avoid the influence of fabric patterns as well as the environment. In the experiments, we proved it works better than the visual method.

(2) A tactile fabric dataset is established, and a visual fabric dataset is established correspondingly. It can be used for subsequent research.

(3) We discuss a novel detection device in the end which can perfectly fit the characteristics of the tactile sensor. It can

*Corresponding author: fangbin@tsinghua.edu.cn.
[1] are from Department of Computer Science and Technology, Tsinghua University, Beijing, China.
[2] is from Department of Automation, Tsinghua University, Beijing, China.

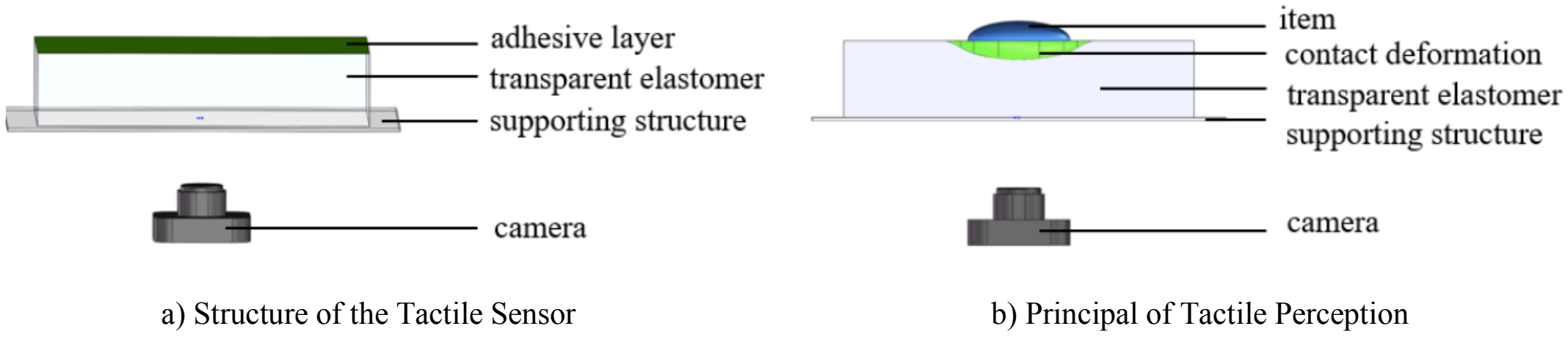

a) Structure of the Tactile Sensor b) Principal of Tactile Perception

**Fig. 1. Visual-based Tactile Sensor**

greatly improve the actual detection efficiency which makes it a possible solution for the actual fabric production.

The paper is organized as follows. Section II briefly reviews the principle of tactile sensors. Section III introduces the dataset, the experimental method, and the detection algorithms. Section IV shows the results of the experiment and the analysis are presented, furthermore discusses the possibility of the tactile detection method in the factory. This paper ends with the conclusion in Section V.

## II. TACTILE SENSOR

Tactile sensation is one of the essential perceptual abilities of humans, which usually can be used as a supplement to vision. The tactile sensation can help us to recognize objects better in different lighting environments, and it also enables us to distinguish objects of different materials with similar appearances, such as similar colors. In order to make the robot's perception ability closer to humans, the tactile sensor is a hot research direction in robotics [28]-[30].

At present, most tactile sensors used for robotic sensing are based on electrical signals, which include resistive sensors [31], capacitive sensors [32], and piezoelectric sensors [33]. The biggest problem with this kind of tactile sensor is that it is difficult to obtain a high tactile resolution, which is not suitable for fabric defect detection.

The vision-based tactile sensor has the characteristics of high resolution, which can get abundant information from the surface of the contact object. Up to now, there are many kinds of research on vision-based tactile sensors, such as Gelforce [34], GelSight [35], and TacTip [36].

According to the requirements of fabric defect detection, we need to be able to detect relatively small structure defects, so we select our tactile sensor with high recognition resolution [37]. As shown in Fig.1a, the main structure of the tactile sensor includes camera, supporting plate, elastomer, adhesive layer and LEDs for internal lighting (not shown in the figure). The camera is installed at the bottom. The elastomer and the supporting plate are both transparent and are fixed directly above the camera. There is an adhesive layer on the surface of the elastomer and its main function is to prevent the influence of ambient light. In addition, the adhesive layer should have good characteristics of reflecting light, so it can help the camera to obtain the tactile information better. Metal powder is usually used as the adhesive layer.

The principle of acquiring tactile information is shown in Fig.1b. When the sensor is in contact with the object, the elastomer will deform to fit the surface texture of the object. The camera then can capture the deformation and get the tactile images of the object's surface.

## III. METHOD

### *A. Dataset*

A total of 10 different kinds of fabrics with different defects are used for data collection. These fabrics contain various common fabric types, patterns and structural defects. For each kind of fabric, we used the tactile sensor to collect about 100 images with defects and about 100 images without defects, obtaining a total of 1966 tactile images. The device we used to collect the tactile dataset is shown in Fig.2.

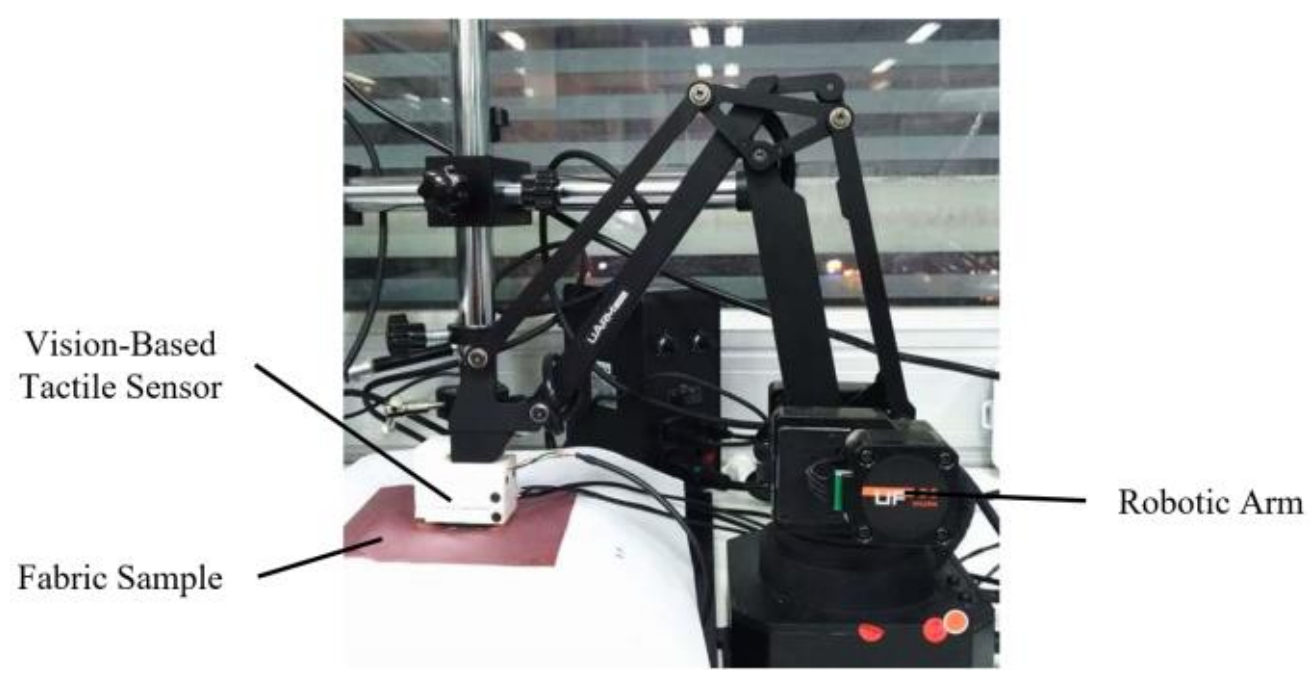

**Fig. 2. Tactile Dataset Collecting System.**

At the same time, for each kind of fabric, we also collected 100 images with defects and 100 images without defects using the camera, a total of 2000 visual images. In order to make the dataset more consistent with the actual use scenarios, the dataset was collected in two different external lighting environments. For example, among the 100 visual images with the defect of fabric type 1, 50 of them were collected in a low lighting environment, and the other 50 were obtained in a high lighting environment. Part of the visual images and their corresponding tactile images are shown in Fig.3. The difference between the visual images and the tactile images can be clearly seen.

The dataset of fabric is shared to the following website, *https://github.com/tsinghua-rll/fabric-texture-dataset*.

### *B. Performance Evaluation*

At present, deep convolutional neural networks are widely used in fabric defect detection. Many good results come from deep learning algorithms [25]-[27]. In order to evaluate the performance of our tactile method, we use a variety of different commonly used convolutional neural network models for the experiment. We train both the tactile dataset and the visual dataset on each of the models to get the detection accuracy, in which way we can compare their performance.

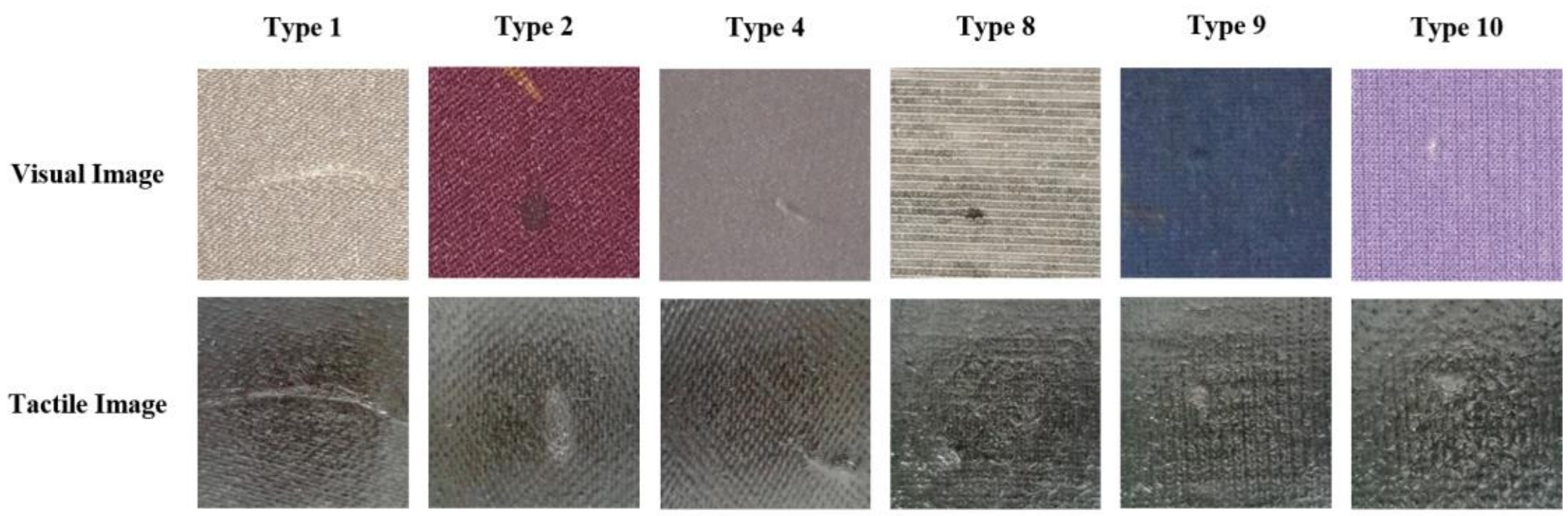

**Fig. 3. Visual Image and Tactile Image.**

The details of the used models will be introduced in the next section.

### C. Efficiency Evaluation

As the fabric defect detection method is proposed for practical use, its efficiency is an important factor that cannot be ignored. In order to measure the detection efficiency, we define the pixel density of fabric detection images. The calculation is as follows.

$$D = N / A \tag{1}$$

where $D$ is the pixel density we defined for a fabric detection image, $N$ is the number of pixels used to display the fabric in the image and $A$ is the actual size of the fabric.

When the algorithm is fixed, the number of image pixels that can be detected per second is certain. Therefore, the lower the pixel density of the fabric detection image, the larger the fabric can be detected at the same time. However, the accuracy of the algorithm will be affected by the decrease in the of pixel density of the image. Based on these assumptions, we conducted an experiment on the accuracy changes of the detection methods under different pixel densites. In the experiment, we changed the resolution to change the number of pixels in the image, so as to change the pixel density. We made a rough conversion for the theoretical detection efficiency based on pixel density, and the conversion formula is as follows.

$$E = f \times n / D \tag{2}$$

where $E$ is the theoretical detection efficiency of the detection method, $f$ is the number of frames detected per second of a certain algorithm and $n$ is the number of pixels for the algorithm's input frame. In this way, we can evaluate the detection efficiency of different methods.

## IV. Experiment & Discussion

### A. Detection Accuracy

We use different deep convolutional neural networks to extract the features of the images and implement the transfer learning method in our training [38]. The structure of our defect detection network is shown in Fig.4.

All of the convolutional neural network models use parameters pre-trained on ImageNet, and we fine-tune the models on our dataset. The number of output features of these models is 1000. We add two fully connected layers after each model to get the output: the input size of the first fully connected layer is 1000 and the output size is 100; the input size of the second fully connected layer is 100 and the output size is 2. The output result is used to judge whether the fabric image has defects.

In the experiment, we use 10-fold cross-validation to get the detection accuracy to ensure the credibility of the results. Each time, we use all the data from 1 type of fabric as the validation set and the data of the other 9 types of fabrics as the training set. At the same time, we also record the results of each training to analyze the difference in detection accuracy between different kinds of fabrics.

The pixel density of fabric detection images used here is 10000 pixels/cm$^2$. Besides, although the use of the RGB color channels can help get more information, the redundant information will make it more difficult for the network to learn the exact features corresponding to the defects, which may cause fluctuation in training. In order to avoid the

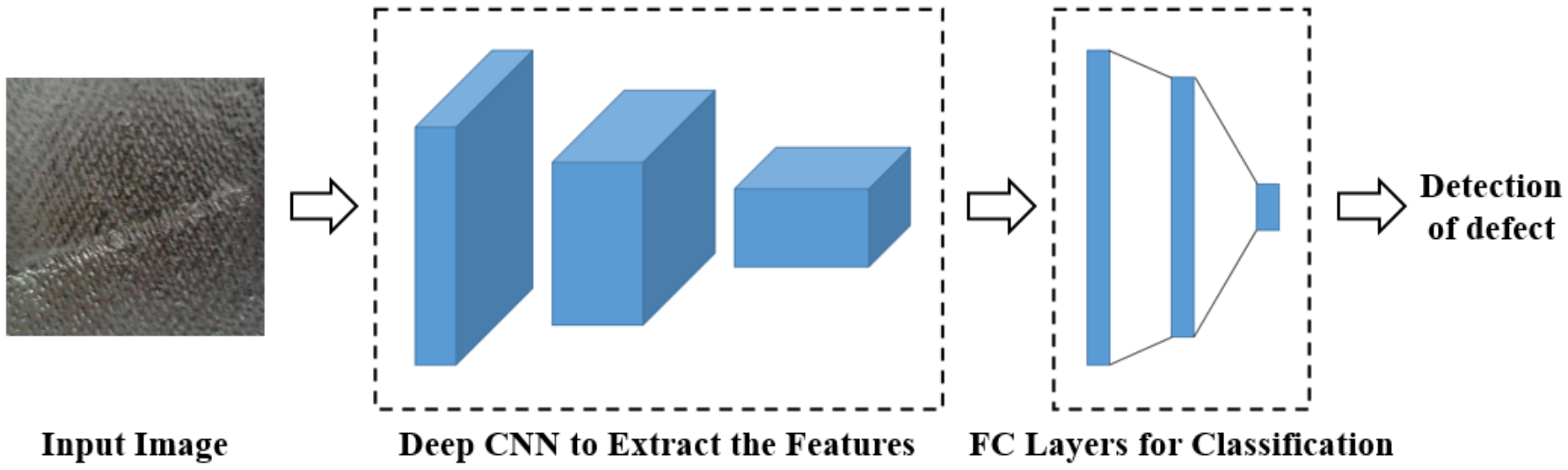

**Fig. 4. Structure of the Defect Detection Network.**

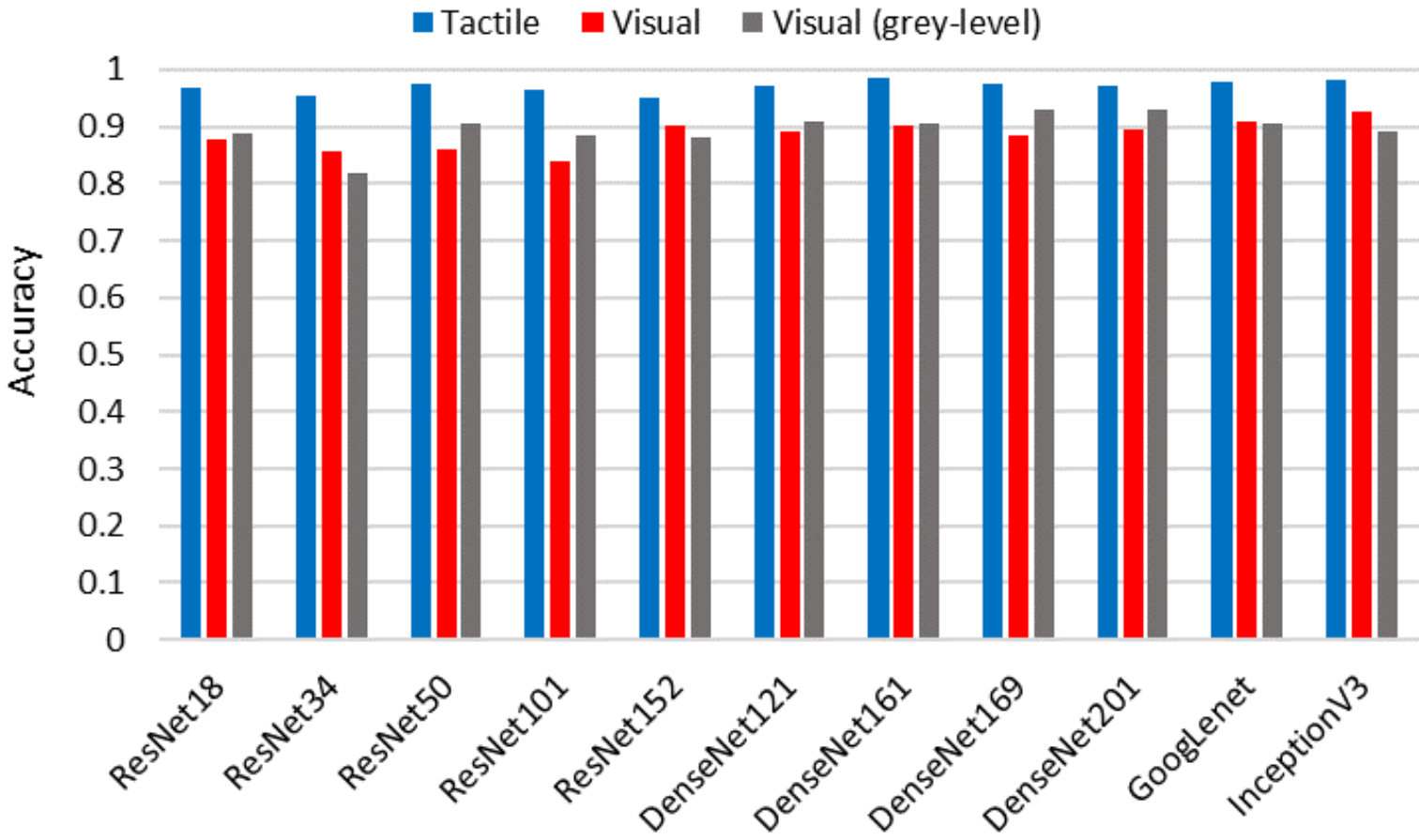

**Fig. 5. Detection Accuracy on Different Models.**

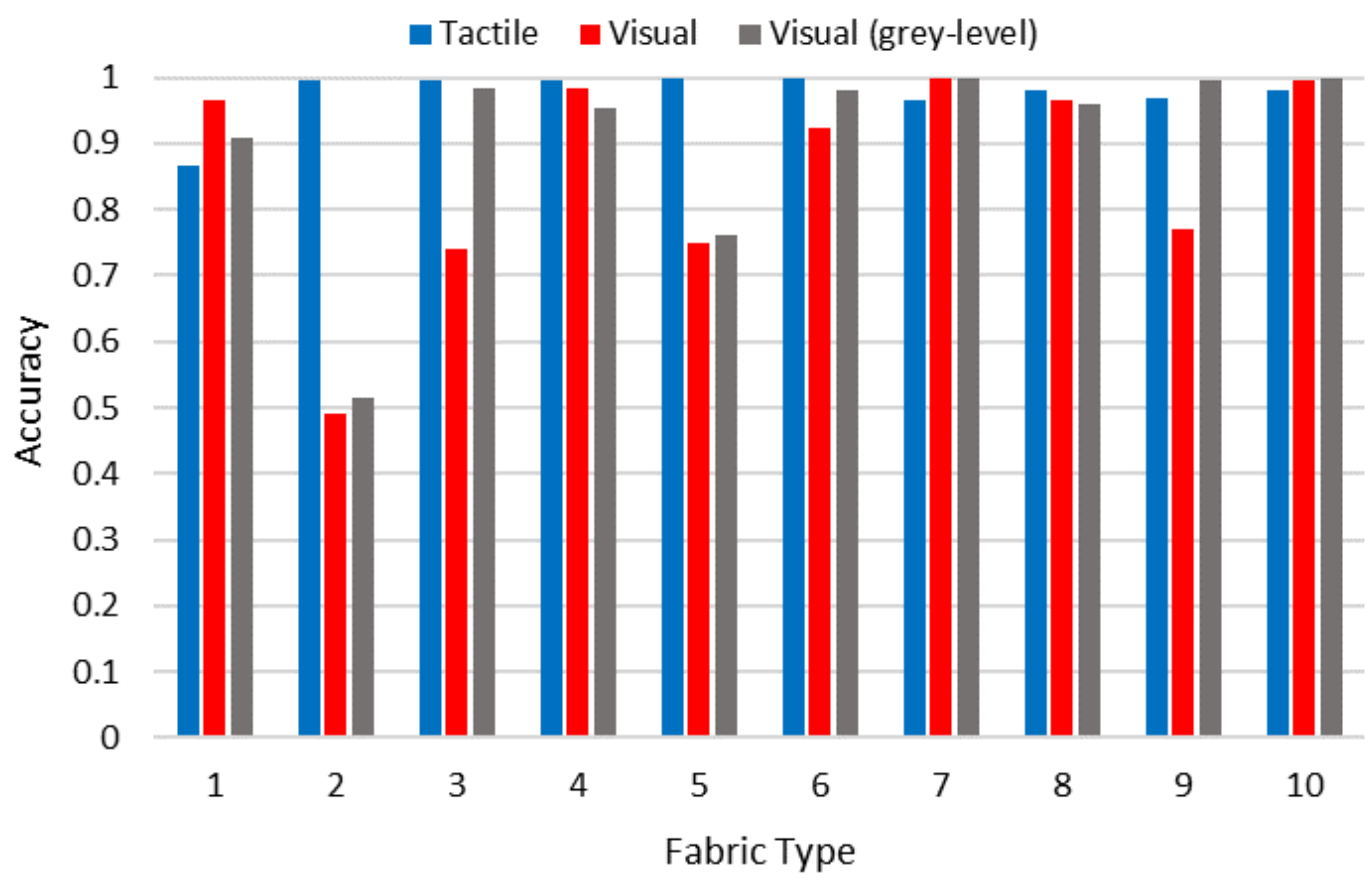

**Fig. 6. Detection Accuracy on Different Fabrics.**

influence of training fluctuation caused by the color channels in visual images, we add gray-level visual images as the control. The results of the experiment are shown in Fig.5:

It can be seen that no matter which deep convolutional neural networks are used to extract the features, the accuracy rate of the method using tactile information is significantly higher than that of the method using visual information. All the detection accuracy results of the tactile method are higher than 95%, and the best accuracy result is 98.36%, which trained on DenseNet161. It can be said that for the fabric structure defects, the method of using tactile information is better than traditional methods of using visual information.

Next, we analyze the results when each type of fabric is used for validation. In this way, we can figure out the specific differences between the two detection methods for different kinds of fabrics, especially when they have different dyeing patterns. We take the results of ResNet50 model, which is most commonly used in image recognition tasks, as an example to analyze.

It can be seen from Fig.6 that the accuracy of the visual method is significantly lower than that of the tactile method when fabric type 2 and type 5 are used as validation sets. Taking fabric type 2 as an example for analysis, the overall image of cloth is shown in Fig.7a. There are irregular dyeing patterns on fabric type 2. The network model trained on the other 9 types of fabric cannot judge whether the pattern is a defect because it has never seen these patterns. The tactile information obtained by the tactile sensor, as shown in Fig.7b, is completely unaffected by these dyeing patterns. Therefore, we can see that tactile information has obvious advantages in the detection of structural defects compared with visual information.

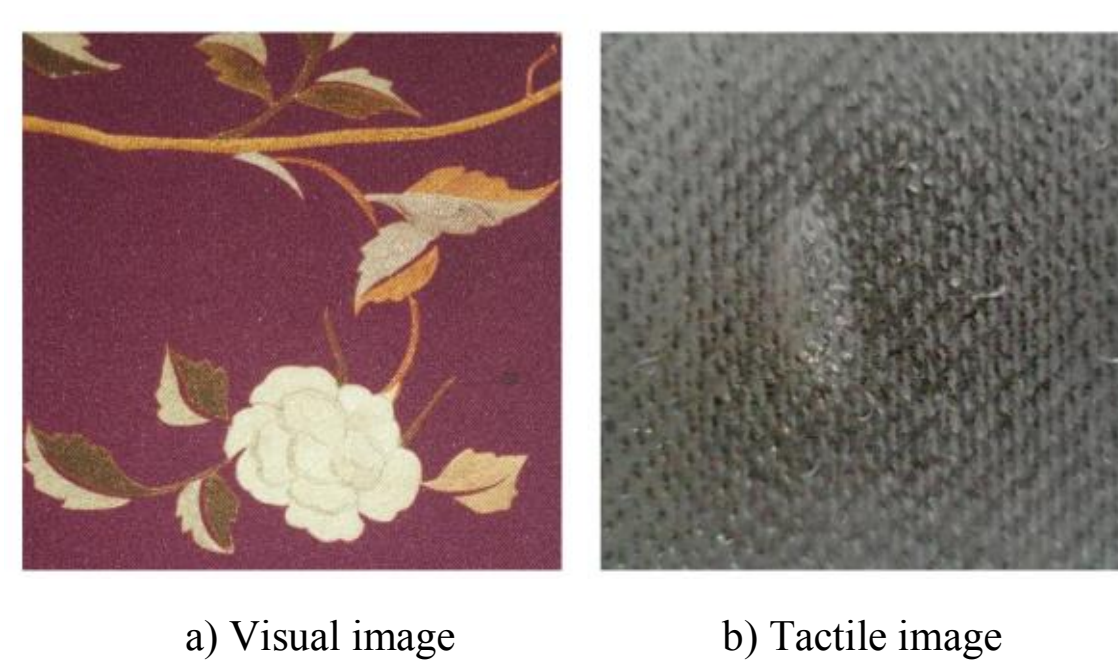

a) Visual image b) Tactile image

**Fig. 7. Visual Image with Irregular Pattern and its Tactile Image.**

## *B. Detection Efficiency*

We still use the most commonly used ResNet50 model to test the fabric detection efficiency. We change the image pixel density by changing the image resolution. The training results of images with different pixel densities are shown in Fig.8. At the same time, we also make a rough conversion for the theoretical detection efficiency according to the pixel density, and the results are also given in Fig.9.

It can be seen that when the pixel density is greater than 1400 pixels/cm$^2$, the detection accuracy can still be maintained at a high level and meet the actual needs. The theoretical detection efficiency for images of this pixel density is 1.55 m$^2$/s, which is enough to meet the requirements of actual fabric production. Therefore, we can draw the conclusion that using tactile information for fabric defect detection can theoretically meet the efficiency requirements, which verifies the feasibility of this method.

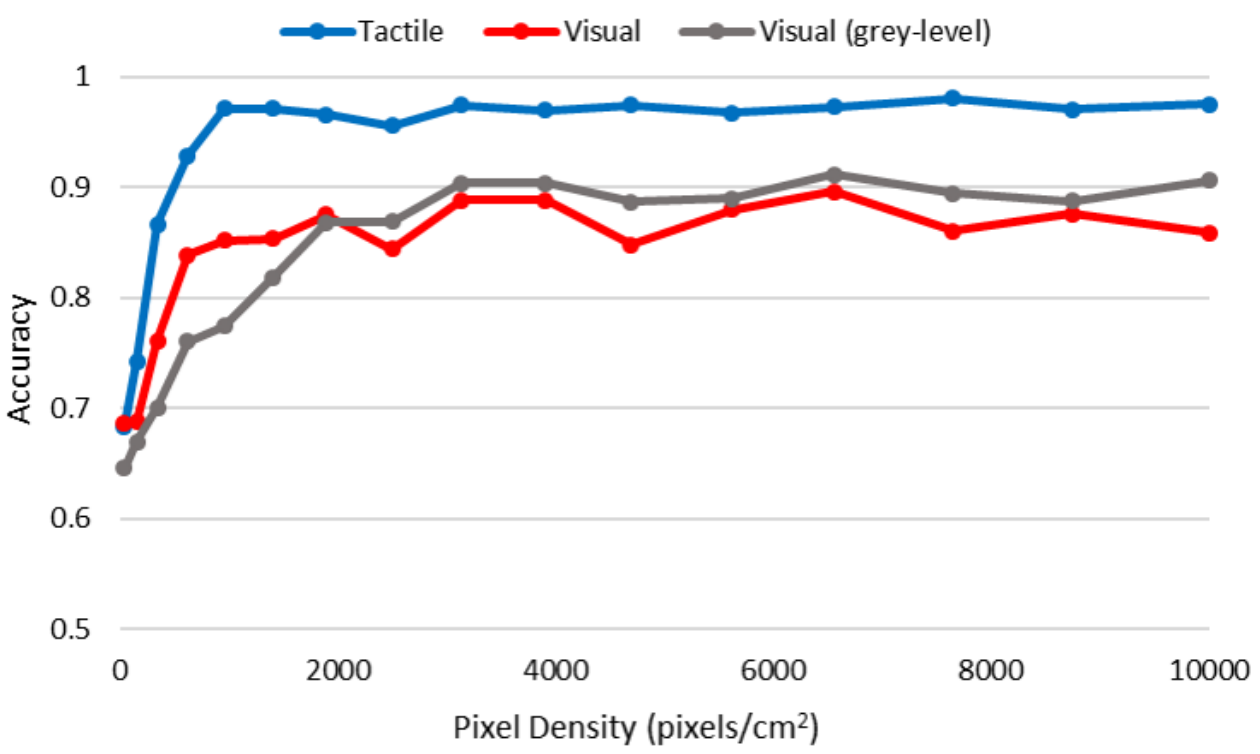

**Fig. 8. Detection Accuracy with Different Pixel Densities.**

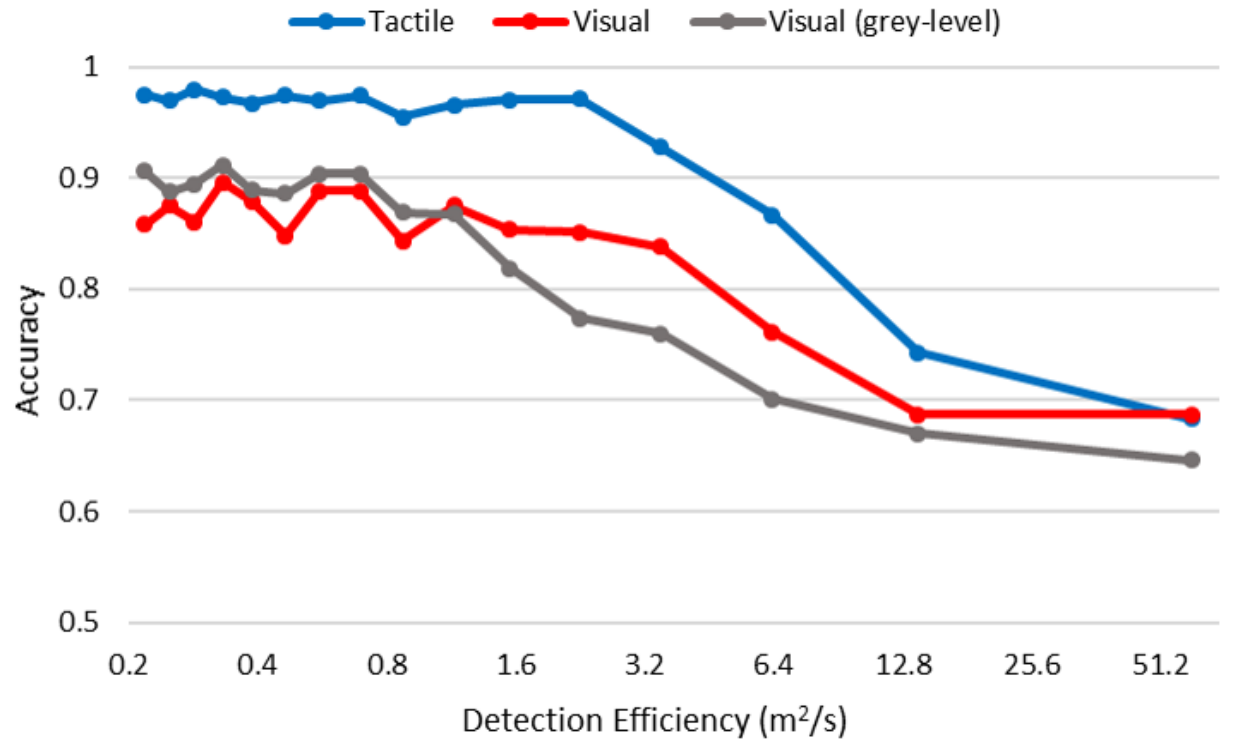

**Fig. 9. Curve of Detection Efficiency.**

## *C. Discussion*

Generally, the sizes of the existing visual-based tactile sensors are usually small. However, in the process of detecting fabric defects, we need to lift, change the position and press the sensor constantly. This process consumes a lot of time, which makes it a time bottleneck. That means the process will make the actual detection efficiency far lower than the theoretical detection efficiency. As a result, we need to find a more efficient tactile detection device to adapt to the fabric production and meet the needs of actual detection efficiency. We then propose a roller type tactile detection device for fabric defect detection, and its structure is shown in the following figure:

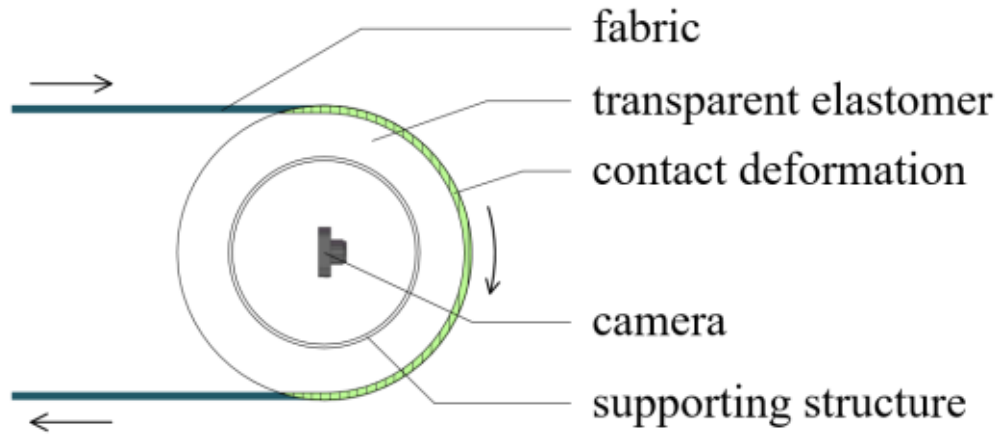

**Fig. 10. Roller Type Tactile Detection Device.**

The composition of the roller-type tactile detection device is similar to the tactile sensor mentioned before, except that the transparent elastomer and the supporting structure are designed in roller shape to adapt to the production of fabrics. The camera is fixed in the center of the device and is always facing the side where the fabric is in contact with the elastomer, recording the deformation.

With the device, the tactile information of the fabric can be obtained constantly without wasting time on moving sensors, so the detection efficiency of the method can be closer to the theoretical detection efficiency now.

## V. Conclusion

We propose a fabric structure defect detection method based on tactile information. The advantage of the method is that it can avoid the problem of different irregular dyeing patterns and reduce the influence of ambient light. Therefore, with the tactile method, the detection algorithm used can be more concise and universal, which makes the system more robust. Besides, we carry out experiments on the tactile method and set the visual method as the comparison group. We get the conclusion that compared with the visual method, the tactile method performs much better in the detection of structural defects. At the same time, we analyze the detection efficiency of the proposed method. The theoretical detection efficiency of the tactile method can meet the demand. In addition, we propose a new tactile sensing device that can make the actual detection efficiency close to the theoretical detection efficiency. In this way, we provide a possible solution for the application of the tactile method in actual fabric production.

The future work will focus on the combination of visual information and tactile information. It can make up for the deficiency that only using tactile information cannot detect color defects. Therefore, we believe that a more suitable industrial fabric detection system can be obtained.

## Acknowledgment

This work was supported by the National Natural Science Foundation of China (Grant No. 91848206), Beijing Science &Technology Project (Grant No. Z191100008019008) and Tsinghua University Department of Computer Science and Technology)-Siemens Ltd., China Joint Research Center for Industrial Intelligence and Internet of Things.

## REFERENCES

[1] A. Kumar, "Computer-Vision-Based Fabric Defect Detection: A Survey," IEEE T. Ind. Electron., vol. 55, no. 1, pp. 348-363, 2008, doi: 10.1109/TIE.1930.896476.

[2] K. Hanbay, M.F. Talu and Ö.F. Özgüven, "Fabric defect detection systems and methods—A systematic literature review," Optik, vol. 127, no. 24, pp. 11960-11973, 2016, doi: https://doi.org/10.1016/j.ijleo.2016.09.110.

[3] H.Y.T. Ngan, G.K.H. Pang and N.H.C. Yung, "Automated fabric defect detection—A review," Image Vision Comput., vol. 29, no. 7, pp. 442-458, 2011, doi: https://doi.org/10.1016/j.imavis.2011.02.002.

[4] M. Alper Selver, V. Avşar and H. Özdemir, "Textural fabric defect detection using statistical texture transformations and gradient search," The Journal of The Textile Institute, vol. 105, no. 9, pp. 998-1007, 2014, doi: 10.1080/00405000.2013.876154.

[5] S. Sadaghiyanfam, "Using gray-level-co-occurrence matrix and wavelet transform for textural fabric defect detection: A comparison study," Proc. 2018 Electric Electronics, Computer Science, Biomedical Engineerings' Meeting (EBBT), 2018, pp. 1-5.

[6] L. Zhoufeng, G. Erjin and L. Chunlei, "A Novel Fabric Defect Detection Scheme Based on Improved Local Binary Pattern Operator," Proc. 2010 International Conference on Intelligent System Design and Engineering Application, 2010, pp. 116-119.

[7] O. Kaynar, Y.E. Işik, Y. Görmez and F. Demirkoparan, "Fabric defect detection with LBP-GLMC," Proc. 2017 International Artificial Intelligence and Data Processing Symposium (IDAP), 2017, pp. 1-5.

[8] H. Bu, J. Wang and X. Huang, "Fabric defect detection based on multiple fractal features and support vector data description," Eng. Appl. Artif. Intel., vol. 22, no. 2, pp. 224-235, 2009, doi: https://doi.org/10.1016/j.engappai.2008.05.006.

[9] M. Hanmandlu, D. Choudhury and S. Dash, "Detection of defects in fabrics using topothesy fractal dimension features," Signal, Image and Video Processing, vol. 9, no. 7, pp. 1521-1530, 2015, doi: 10.1007/s11760-013-0604-5.

[10] K. Sakhare, A. Kulkarni, M. Kumbhakarn and N. Kare, "Spectral and spatial domain approach for fabric defect detection and classification," Proc. 2015 International Conference on Industrial Instrumentation and Control (ICIC), 2015, pp. 640-644.

[11] L. Yihong and Z. Xiaoyi, "Fabric Defect Detection with Optimal Gabor Wavelet Based on Radon," Proc. 2020 IEEE International Conference on Power, Intelligent Computing and Systems (ICPICS), 2020, pp. 788-793.

[12] Y. Li and X. Di, "Fabric defect detection using wavelet decomposition," Proc. 2013 3rd International Conference on Consumer Electronics, Communications and Networks, 2013, pp. 308-311.

[13] V.V. Karlekar, M.S. Biradar and K.B. Bhangale, "Fabric Defect Detection Using Wavelet Filter," Proc. 2015 International Conference on Computing Communication Control and Automation, 2015, pp. 712-715.

[14] A. Dogandžić, N. Eua Anant and B. Zhang, "Defect Detection Using Hidden Markov Random Fields," AIP Conference Proceedings, vol. 760, no. 1, pp. 704-711, 2005, doi: 10.1063/1.1916744.

[15] C. Zhang, J. Liu, C. Liang, Z. Xue, J. Pang and Q. Huang, "Image classification by non-negative sparse coding, correlation constrained low-rank and sparse decomposition," Comput. Vis. Image Und., vol. 123, pp. 14-22, 2014, doi: https://doi.org/10.1016/j.cviu.2014.02.013.

[16] B. Shi, J. Liang, L. Di, C. Chen and Z. Hou, "Fabric defect detection via low-rank decomposition with gradient information and structured graph algorithm," Inform. Sciences, vol. 546, pp. 608-626, 2021, doi: https://doi.org/10.1016/j.ins.2020.08.100.

[17] D. Mo, W.K. Wong, Z. Lai and J. Zhou, "Weighted Double-Low-Rank Decomposition with Application to Fabric Defect Detection," IEEE T. Autom. Sci. Eng., pp. 1-21, 2020, doi: 10.1109/TASE.2020.2997718.

[18] H. Liu, I. Chatterjee, M. Zhou, X. S Lu and A. Abusorrah, "Aspect-Based Sentiment Analysis: A Survey of Deep Learning Methods," IEEE Transactions on Computational Social Systems, vol. 7, no. 6, pp. 1358-1375, 2020, doi: 10.1109/TCSS.2020.3033302.

[19] B. Fang, X. Ma, J. Wang, F. Sun, "Vision-based posture-consistent teleoperation of robotic arm using multi-stage deep neural network", Robotics Auton. Syst., 2020, 131: 103592.

[20] B. Fang, Q. Zhou, F. Sun, J. Shan, M. Wang, C. Xiang, Q. Zhang, "Gait Neural Network for Human-Exoskeleton Interaction", Frontiers Neurorobotics, 2020, 14: 58.

[21] C. Szegedy, et al., "Going deeper with convolutions", Proceedings of the IEEE Conference on Computer Vision and Pattern Recognition (CVPR), 2015, pp. 1-9.

[22] C. Szegedy, V. Vanhoucke, S. Ioffe, J. Shlens and Z.B. Wojna, "Rethinking the Inception Architecture for Computer Vision", Proceedings of the IEEE Conference on Computer Vision and Pattern Recognition (CVPR), 2016, pp. 2818-2826.

[23] K. He, X. Zhang, S. Ren and J. Sun, "Deep Residual Learning for Image Recognition", Proceedings of the IEEE Conference on Computer Vision and Pattern Recognition (CVPR), 2016, pp. 770-778.

[24] G. Huang, Z. Liu, L. van der Maaten and K. Weinberger, "Densely Connected Convolutional Networks", Proceedings of the IEEE Conference on Computer Vision and Pattern Recognition (CVPR), 2017, pp. 4700-4708.

[25] Z. Liu, B. Wang, C. Li, B. Li and X. Liu, "Fabric Defect Detection Algorithm Based on Convolution Neural Network and Low-Rank Representation," Proc. 2017 4th IAPR Asian Conference on Pattern Recognition (ACPR), 2017, pp. 465-470.

[26] A. Şeker, "Evaluation of Fabric Defect Detection Based on Transfer Learning with Pre-trained AlexNet," Proc. 2018 International Conference on Artificial Intelligence and Data Processing (IDAP), 2018, pp. 1-4.

[27] W. Ouyang, B. Xu, J. Hou and X. Yuan, "Fabric Defect Detection Using Activation Layer Embedded Convolutional Neural Network," IEEE Access, vol. 7, pp. 70130-70140, 2019, doi: 10.1109/ACCESS.2019.2913620.

[28] H. Yousef, M. Boukallel and K. Althoefer, "Tactile sensing for dexterous in-hand manipulation in robotics—A review," Sensors and Actuators A: Physical, vol. 167, no. 2, pp. 171-187, 2011, doi: https://doi.org/10.1016/j.sna.2011.02.038.

[29] Z. Kappassov, J. Corrales and V. Perdereau, "Tactile sensing in dexterous robot hands — Review," Robot. Auton. Syst., vol. 74, pp. 195-220, 2015, doi: https://doi.org/10.1016/j.robot.2015.07.015.

[30] F. Sun, B. Fang, etal, "A novel multi-modal tactile sensor design using thermochromic material," Sci. China Inf. Sci., 2019, 62(11): 214201:1-214201:3.

[31] S. Stassi, C. Valentina, G. Canavese and C. Pirri, "Flexible Tactile Sensing Based on Piezoresistive Composites: A Review," Sensors (Basel, Switzerland), vol. 14, pp. 5296-5332, 2014, doi: 10.3390/s140305296.

[32] P. Maiolino, M. Maggiali, G. Cannata, G. Metta and L. Natale, "A Flexible and Robust Large Scale Capacitive Tactile System for Robots," IEEE Sens. J., vol. 13, no. 10, pp. 3910-3917, 2013, doi: 10.1109/JSEN.2013.2258149.

[33] C. Chuang, M. Wang, Y. Yu, C. Mu, K. Lu and C. Lin, "Flexible tactile sensor for the grasping control of robot fingers," Proc. 2013 International Conference on Advanced Robotics and Intelligent Systems, 2013, pp. 141-146.

[34] K. Kamiyama, H. Kajimoto, M. Inami, N. Kawakami and S. Tachi, "Development of A Vision-based Tactile Sensor," IEEJ Transactions on Sensors and Micromachines, vol. 123, pp. 16-22, 2001, doi: 10.1541/ieejsmas.123.16.

[35] M.K. Johnson and E.H. Adelson, "Retrographic sensing for the measurement of surface texture and shape," Proc. 2009 IEEE Conference on Computer Vision and Pattern Recognition, 2009, pp. 1070-1077.

[36] C. Chorley, C. Melhuish, T. Pipe and J. Rossiter, "Development of a tactile sensor based on biologically inspired edge encoding," Proc. 2009 International Conference on Advanced Robotics, 2009, pp. 1-6.

[37] B. Fang, F. Sun, H. Liu, "A dual-modal vision-based tactile sensor for robotic hand grasping", International Conference on Robotics and Automation, 2018, 2483-2488

[38] J. Yosinski, J. Clune, Y. Bengio and H. Lipson, "How transferable are features in deep neural networks?" Advances in Neural Information Processing Systems (NIPS), vol. 27, 2014.